\documentclass{article}

\usepackage{arxiv}

\usepackage[utf8]{inputenc} 
\usepackage[T1]{fontenc}    
\usepackage{hyperref}       
\usepackage{url}            
\usepackage{booktabs}       
\usepackage{amsfonts}       
\usepackage{nicefrac}       
\usepackage{microtype}      
\usepackage{lipsum}
\usepackage{graphicx}
\graphicspath{ {./images/} }

\usepackage{microtype}
\usepackage{graphicx}
\usepackage{booktabs} 
\usepackage{amsmath, amsfonts}
\usepackage{mathtools}
\usepackage{hyperref}

\usepackage{subcaption}
\newcommand{\kl}[2]{D_{\rm KL}(#1 \ \| \ #2)}

\usepackage[]{algorithm}
\usepackage[noend]{algorithmic}
\newcommand{\model}{\mathsf{NP}^{\mathbf{w}}}

\title{Improved Memories Learning}

\author{
Francesco Varoli \\
Computational Science and Engineering Laboratory\\
ETH Zürich\\
Switzerland, CH-8092 \\
\texttt{varolif@student.ethz.ch} \\
\And
Guido Novati \\
Computational Science and Engineering Laboratory\\
ETH Zürich\\
Switzerland, CH-8092 \\
\texttt{novatig@ethz.ch} \\
\And
Pantelis R.~Vlachas \\
Computational Science and Engineering Laboratory\\
ETH Zürich\\
Switzerland, CH-8092 \\
\texttt{pvlachas@ethz.ch} \\
\And
Petros Koumoutsakos \\
Computational Science and Engineering Laboratory\\
ETH Zürich\\
Switzerland, CH-8092 \\
\texttt{petros@ethz.ch} \\
}

\begin{document}
\maketitle
\begin{abstract}
We propose \textit{Improved Memories Learning (IMeL)}, a novel algorithm that turns reinforcement learning (RL) into a supervised learning (SL) problem and delimits the role of neural networks (NN) to interpolation.
IMeL consists of two components.
The first is a reservoir of experiences.
Each experience is updated based on a non-parametric procedural improvement of the policy, computed as a bounded one-sample Monte Carlo estimate.
The second is a NN regressor,
which receives as input improved experiences from the reservoir (context points)
and computes the policy by interpolation.
The NN learns to measure the similarity between states in order to compute long-term forecasts by averaging experiences, rather than by encoding the problem structure in the NN parameters.
We present preliminary results and propose IMeL as a baseline method for assessing the merits of more complex models and inductive biases.
\end{abstract}

\section{Introduction}
Deep reinforcement learning (RL) algorithms have produced many of the seminal results of machine learning, achieving success in classic games \cite{silver2017mastering}, videogames \cite{mnih2015human,vinyals2019grandmaster}, and robotic control \cite{andrychowicz2020learning}.
These results required extraordinary feats in terms of algorithmic and modeling advances, as well as high-performance computing to produce the large amounts of data required for training. Along with these advances, the research credit assignment problem (that of isolating the causes of the performance gains) grows in complexity.
Indeed, recent works have shown that the results obtained by deep RL rely, in unintuitive ways, on code-level optimizations \cite{engstrom2019implementation}, hyper-parameter tuning \cite{henderson2018deep}, and unreliable gradient estimates \cite{ilyas2018deep}.
Here we focus on the question ``are deep neural networks (NNs) beneficial to RL because they are able to encode a model of the problem?'' and attempt the investigation by contradiction.

RL algorithms can be broadly classified as either model-based or model-free. Model-based methods explicitly learn a model of the environment to predict and plan future steps, while the model-free ones define objective functions that can be computed directly from environment-interaction data.
In the case of deep RL, many model-free methods learn value functions which implicitly learn the structure of the environment by constructing an utilitarian model to predict future outcomes.
For the purpose of efficiently adapting to specific tasks, inductive biases may be presumed to benefit both model-based and model-free methods, as they allow encoding prior knowledge about the RL problem in the parametric architecture.
A common pitfall encountered by RL methods is that of modeling ``delusions'': structural mis-representations of the task encoded in the NN models.
For example, one such delusion was faulted for the loss of AlphaGo on the fourth game against Lee Sedol, while \cite{lu2018non} show that incorrect inductive biases (such as linear approximation for Q-learning) are a cause for delusional bias.
Structural errors in model-based RL have been shown to decrease stability \cite{van2019use} and generalization \cite{zhang2018dissection} compared to methods that rely purely on experience replay.
Despite the fact that incorrect inductive biases are detrimental for RL, NN architectures commonly employed for RL are those employed also for supervised learning rather than specifically optimized for RL.
Moreover, linear approximation provides a strong baseline for policy optimization in continuous control benchmarks \cite{rajeswaran2017towards}.

In order to asses whether (implicit or explicit) RL models can surpass methods that rely exclusively on experiences in a replay memory (RM), we propose a novel baseline algorithm, \textit{Improved Memories Learning (IMeL)}.
IMeL does not rely on a NN to predict value functions or compute optimal actions.
Instead, the NN is employed to interpolate between experiences and is trained with a supervised learning loss.
The IMeL agent, similarly to many off-policy RL methods, fills and maintains a reservoir of experiences. IMeL performs policy optimization by annotating each experience with an estimate improved policy, computed as an incremental Monte Carlo update of the behavior policy, as described in Sec.~\ref{sec:method}. Therefore, by construction the IMeL algorithm is less reliant on the ability of NN to learn the structure of the task and generalize to unseen states, rather the NN serves only for regression. We claim that a properly tuned implementation of IMeL may serve as a simple baseline against which the benefits of models and inductive biases can be fairly measured.

\section{Background and Related Work}
\label{sec:related_work}
The IMeL algorithm takes inspiration from several well-known practices in RL.
The policy gradient theorem~\cite{sutton2000}, trust region updates~\cite{schulman2015trust} and Experience Replay~\cite{lin1992self}
are the building blocks constituting IMeL.
Here we summarize these concepts to provide the framework in which the algorithm is developed.

\subsection{Constrained Policy Optimization}
Policy gradient methods directly optimize the policy by searching the class of functions $\pi_{\theta}(a|s)$ parametrized by $\theta$.
The search is performed by gradient ascent, maximizing the performance measure $J(\theta) = \mathbb{E}_{s,a}\big[\sum_{t=0}^T{\gamma^t r_t}\big]$ by following the policy gradient:
\begin{equation}
    \nabla_{\theta}J(\theta)= \mathbb{E}_{\pi_{\theta}}\Big{[} Q_{\pi_{\theta}}(s, a) \ \nabla_{\theta}   \log \pi_{\theta}(a|s)  \Big{]}
\end{equation}
Estimations of this gradient are known to have high variance and depend on the choice of estimator for the on-policy state-action value $Q_{\pi_{\theta}}(s, a)$~\cite{sutton1998introduction}. The state-action value is usually approximated by a parametric model, computed with one-sample Monte Carlo (MC) estimates, or combinations thereof. These sources of inaccuracy about the gradient's \emph{direction} are compounded with the uncertainty about the correct step \emph{size}~\cite{Kakade02approximatelyoptimal}.
The most prominent way to stabilize policy gradient-based algorithms is that of limiting each step's optimization search space by employing constrains, as in Trust Region Policy Optimization (TRPO) \cite{schulman2015trust}, or penalization, as in Proximal Policy Optimization (PPO) \cite{schulman2017proximal}.
Here we both use as baseline and adapt the techniques of TRPO. 
On each step, TRPO solves an optimization problem for $\theta$ constrained on the average Kullback-Leibler (KL) divergence $D_{KL}\left(\pi_{\theta_{old}} \ \| \ \pi_{\theta} \right)$. Here $\pi_{\theta_{old}}$ is the policy after the previous step which is used to sample the environment and to compute the MC estimates of the Q function for the current step, i.e. $q_{\pi_{\theta_{old}}}(s, a)$.
The optimization problem can be written as:
\begin{align}
\max_{\theta} ~& 
\mathbb{E} \left[ \left. \frac{\pi_{\theta}(a|s)}{\pi_{\theta_{old}} (a|s)} q_{\pi_{\theta_{old}}}(s, a)~\right|~\substack{s \sim \rho_{\theta_{old}} \\ a \sim \pi_{\theta_{old}} }\right]
\label{eq:opt:1}\\
\text{s.t.} ~&
\mathbb{E} \left[ \left.
\kl{\pi_{{\theta}_{old}}(\cdot|s)}
{\pi_{\theta}(\cdot|s)} ~
\right| ~ s \sim \rho_{\theta_{old}}
\right]
\le \delta
\end{align}
Here $\rho_{\theta_{old}}$ is the state visitation frequency induced by the policy $\pi_{{\theta}_{old}}$.
TRPO efficiently iterates this optimization problem by linear approximation of the objective and quadratic approximation of the constraint equation. TRPO is a simple and stable algorithm and has become a benchmark for several continuous problems.

\subsection{Experience Replay}
Experience Replay (ER) has been widely employed for developing sample efficient off-policy RL algorithms.
ER allows the algorithm to reuse experiences over multiple updates, boosting its sample efficiency \cite{lin1992self}.
For each environment time-step $t$, experiences can be stored in a replay memory (RM) as tuples $\{s_t, r_t, a_t, \beta_t\}$, where $s_t$ is the state, $r_t$ is the reward, and $\beta_t$ represents the policy statistics (mean and covariance for continuous actions, categorical probabilities for discrete actions) used to sample the action $a_t$.
However, as the policy is trained it becomes increasingly different from past behaviours. Even with off-policy RL methods, updates computed using distant experiences become inaccurate and may lead to instabilities \cite{novati2018remember}.

\subsection{Supervised Reinforcement Learning}
RL approaches typically attempt to infer actions that correspond to maximal rewards or train function approximators to satisfy the Bellman equation. Until recently, the use of Supervised Learning (SL) methods in RL were limited to, for example, learning models of the environment's dynamics or of the reward function.
A Recently proposed alternative, termed \textit{Upside-Down Reinforcement Learning} (UDRL) \cite{schmidhuber2019reinforcement, srivastava2019training},
is to train a function approximator to predict actions that correspond to a desired reward and time horizon (\textit{behaviour function}).
Because the NN takes the returns as input, rather than having to predict them, and outputs the corresponding action, it can be trained with a SL loss by experience replay.
UDRL obtained competitive results in several RL problems.

\section{Supervised learning methods}
\label{sec:sl_methods}
The proposed IMeL algorithm requires NNs only as interpolators among past experiences.
Therefore, we employ NN architectures designed with inductive biases of interpolation which are trained by SL. In this section we present the methods used for this task.

\subsection{Neural Processes}

Neural Processes (NP) \cite{garnelo2018neural} are generative models able to model a distribution over functions, combining the benefits of NNs and Gaussian Processes.
NPs model a random process and learn this distribution in a data-driven manner.
This is achieved by introducing a global latent variable $z$, assumed to be modeling the randomness of the process, and a nonlinear regression function $g(z, x)$.
NPs learns the distribution of $z$ by an encoder network $h^{\mathbf{w}}(\mathcal{C})$ mapping the set of available observations $\mathcal{C} = \{ (x,y)_i \}_{i=1}^m$ to the parameters $\mu_z, \sigma_z$ defining the conditional distribution $p(z|\mathcal{C}) = \mathcal{N}(\mu_z, \sigma_z|\mathcal{C})$.
A decoder network $g^{\mathbf{w}}(z, x)$ is mapping the latent variable realization $\overline{z} \sim p(z|\mathcal{C})$ and the target set $x_{1:n}$ to the mean and standard deviation of the posterior distribution $y_{1:n} \sim \mathcal{N}(\mu_y, \sigma_y| x_{1:n}, \mathcal{C})$.
To learn such mappings, NPs employ amortised variational inference, minimizing the cost $\mathcal{L}$ derived by the evidence lower-bound (ELBO), i.e.
\begin{equation}
\begin{aligned}
\mathcal{L} =
\mathbb{E}_{
q(z|x_{1:n}, y_{1:n})
}
\Bigg[
\sum_{i=m+1}^{n} &-\log{p(y_i|x_i, z)}
\\
&+\log{\frac{q(z|x_{1:n}, y_{1:n})}{q(z|x_{1:m}, y_{1:m})}} \Bigg]
\label{eq:L}
\end{aligned}
\end{equation}
where the distributions $q(z|x_{1:n}, y_{1:n})$ and $q(z|x_{1:m}, y_{1:m})$ are the variational posteriors computed with the available context set.
The first term in Equation~\ref{eq:L} maximizes the log-likelihood of the samples given $z$ and is thus used to learn $g^{\mathbf{w}}(x, z)$.
The second term is the Kullback-Leibler divergence between the prior $z$ distribution, conditioned on the context set alone, and the posterior one which considers the target set as well. 
Minimizing the kl divergence, the algorithm learns how to approximate the distribution of the latent variable from the context set.
The posterior distribution $y_{1:n} \sim \mathcal{N}(\mu_y, \sigma_y | x_{1:n}, \mathcal{C})$ gives an estimate of the uncertainty in the prediction, which largely depends on the amount of the nearby context points.
Eventually, NPs manage to learn a distribution over functions induced by the realizations $\overline{z}$ of $z$.
The trained model is employed for probabilistic inference.

\subsection{Mean Kernel Interpolation}

Mean Kernel Interpolation (MKI) is a simple architecture explicitly designed to perform weighted interpolation of a set of experiences. 
The model is constituted by two parts.
A NN $z=f^{\mathbf{w}}(x)$ projecting the input $x$ to a feature space $z$, and an exponential kernel function parametrized by $\mathbf{W}$ used to represent the geometric mapping between the latent space $z$ and the output space $y$.
Interpolation on the $y$ space is performed by interpolating the kernel in the latent space, based on the context set points.

By training the architecture (learning the weights $\mathbf{w}$ and $\mathbf{W}$) on a dataset of functions, the NN learns to extract the common features to capture the global structure of the function space.
To perform inference on a target point $x_{n} \in \mathbb{R}^{d_x}$ given the context set $\mathcal{C} = \{(x, y)_i\}_{i=1}^{m}$, we first project the x-values onto the feature space $z \in \mathbb{R}^{d_z}, \ d_z < d_x$.
Then the y-values of the context points are interpolated using the learned exponential kernel, parametrized by the weight matrix $\mathbf{W} \in \mathbb{R}^{d_z \times d_z}$, i.e.
\begin{equation}
    y_n = \sum_{i=1}^m y_i \exp{ \big (
    (z_n-z_i)
    \mathbf{W}
    (z_n-z_i)
    \big)}
    ,
\end{equation}
where $z_n$ is the projection of the target point, while $\{z_i\}_{i=1}^m$ the projection of the context set.
This procedure results in a method able to efficiently interpolate the context points and accurately regress the function. As opposed to NPs, MKI is not a generative model.

\section{IMeL algorithm}
\label{sec:imel_alg}

\begin{figure}[h!]
\centering
\includegraphics[trim=.8cm 1.cm .3cm .6cm,clip=true,width=.85\linewidth]{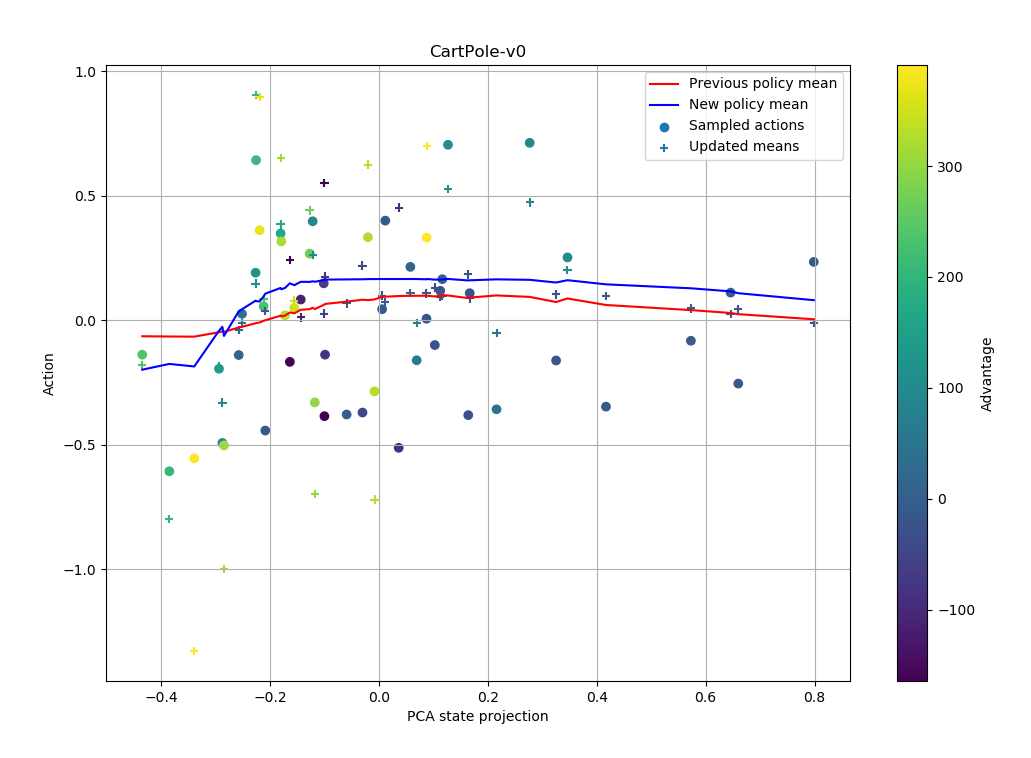}
\caption{
Illustration of the update step of IMeL in the Cartpole problem.
The NN interpolation of the improved means into the new policy is plotted against the PCA projection of the four-dimensional state space.
}
\label{fig:up_sim}
\end{figure}

As in standard policy gradient methods, we optimize a parametrized policy with respect to its expected discounted reward.
However, instead of directly updating its parameters, we
\textbf{(1)} update previously sampled policy realizations,
\textbf{(2)}
utilize a supervised learning method to interpolate on the space of policies to
\textbf{(3)}
obtain the new policy function.
To ensure robustness, taking inspiration from TRPO, we constrain the change between the sampled mean-distribution and the improved one to be within a trust region.
All the improved policy realizations, are stored in a replay memory,
so that the SL method can keep on learning from them at later iterations.
Learning from this \textbf{distribution of improved policies},
instead of directly updating the parameters, allows to best exploit the local information contained in the dataset and use the information of each experience where most needed.
An update step of IMeL is illustrated in Figure~\ref{fig:up_sim}.

\subsection{The structure}\label{sec:method}
For this pilot version of the algorithm we consider only continuous-control problems and Gaussian policies $\pi(a|s)$.
We augment every experience $\phi_t = (s_t, \mu_t, \sigma_t, a_t, r_t)$ gathered during the interaction with the environment with an improved mean of the policy $\mu_t^*$, which can be thought of as a free parameter.
The gradient of the Monte Carlo return with respect to $\mu_t$ is estimated by the policy gradient:
\begin{equation*}
\begin{split}
    \nabla_{\mu_t} J &= \nabla_{\mu_t} \big[ \log ( \pi(a_t|s_t) ) \ q_{\pi}(s_t, a_t)\big]\\ &\approx \nabla_{\mu_t} \Big{[} \log \Big{(} \frac{1}{\sigma_t \sqrt{2 \pi}} e^{-\frac{(a_t-\mu_t)^2}{2\sigma_t^2}}\Big{)} \Big{]} q_{\pi}(s_t, a_t) \\
    &=   \Big{(}\frac{a_t-\mu_t}{\sigma_t^2}\Big{)}  q_{\pi}(s_t, a_t)
    .
\end{split}
\end{equation*}
Here, $q_{\pi}(s_t, a_t)$ is the Monte Carlo return.
This result is used to generate the improved policy $\mu_t^*$, i.e.
\begin{equation*}
\begin{split}
    \mu_t^* &= \mu_t
    + \eta \nabla_{\mu_t}J
    \approx \mu_t + \eta \ q_{\pi}(s_t, a_t)  \Big{(}\frac{a_t-\mu_t}{\sigma_t^2}\Big{)}
    ,
\end{split}
\end{equation*}
The step size $\eta$ is a crucial parameter, since it determines how far the improved means will be moved away from the sampled ones, tuning the trade-off between accuracy and learning speed.
Taking inspiration from TRPO, we take $\eta$ such that the expected KL divergence between the behavior policy and the improved policies is bounded by a hyper-parameter $\epsilon$.
Assuming a constant standard deviation over the iterations $\sigma$ and using MC sampling to estimate the mean $D_{KL}$ over the state space, we obtain the following analytical relations:  
\begin{equation*}
\epsilon =
    \mathbb{E}_{s}\Big[D_{KL}\Big(\pi^*(a_t|s_t) || \pi(a_t|s_t) \Big)\Big] \approx
    \frac{1}{T}\sum_{t=0}^T \frac{(\mu_t^*- \mu_t)^2}{2 \sigma_t^2}
    .
\end{equation*}

Substituting the update step relation $ \mu_t^* =  \mu_t + \eta \ q_t^{\pi} \ \frac{a_t - \mu_t}{\sigma^2}$
we obtain: 
\begin{equation}
\begin{split}
\epsilon &  =
    \frac{1}{T} \sum_{t=0}^T \frac{(\mu_t + \eta \ q_t^{\pi} \ \frac{a_t - \mu_t}{\sigma^2} - \mu_t)^2}{2 \sigma_t^2}
    \implies
    \\
    \epsilon & =
    \frac{1}{T} \ \eta^2 \ {q_t^{\pi}}^2 \sum_{t=0}^T \frac{(a_t - \mu_t)^2}{2\sigma_t^6}
    \implies
    \\
    \eta & = \sqrt{\frac{2 \ T \ \epsilon}{{q_t^{\pi}}^2 \sum_{t=0}^T \frac{(a_t - \mu_t)^2}{2\sigma_t^6} }}
    .
\end{split}
\end{equation}
Therefore, the RL process of estimating optimal policies given states is delegated to an analytical procedure, rather than relying on function approximation. Yet, these estimates are noisy and imprecise due to the Monte Carlo estimate of the expected returns.
A similar procedure may be employed to update the standard deviation, which we kept constant in this study, or could be derived for discrete policies.

By moving all experiences in the Replay Memory towards the direction of higher returns, IMeL generates a dataset of improved policies $\mathcal{C} = \{(s, \mu^*)_i\}_{i=1}^m$.
For an unseen state $s$, the policy $\pi(a|s)$ is computed by interpolation given a learned kernel (via NPs and MKI) which determines the similarity of $s$ to the set of states in $\mathcal{C}$ and all prior improved policies.
The NN regressor is trained with a SL loss to predict $\mu^*$ for a state excluded from the set $\mathcal{C}$ by interpolating between the noisily improved experiences in the RM.
The pseudocode for the IMeL algorithm is reported in the Supplementary Material.

\begin{figure}[h!]
\centering
\begin{minipage}{0.5\linewidth}
\includegraphics[trim=.5cm 0cm .6cm .6cm,clip=true,width=\linewidth]{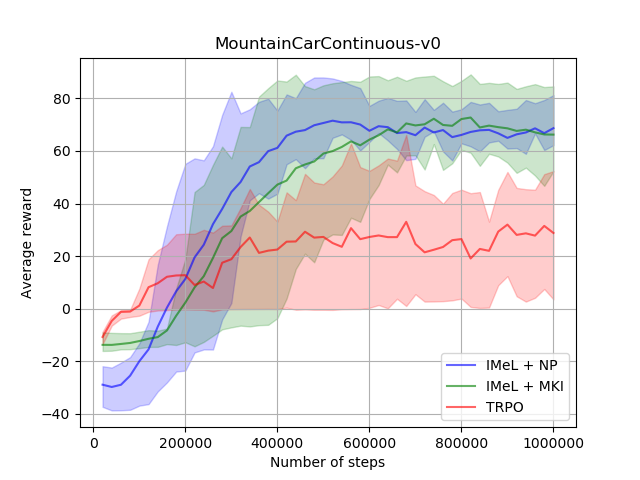}
\centering
\subcaption{
$s \in \mathbb{R}^2$,
$a \in \mathbb{R}$
}
\end{minipage}%
\hfill
\centering
\begin{minipage}{0.5\linewidth}
\includegraphics[trim=.8cm 0cm .3cm .6cm,clip=true, width=\linewidth]{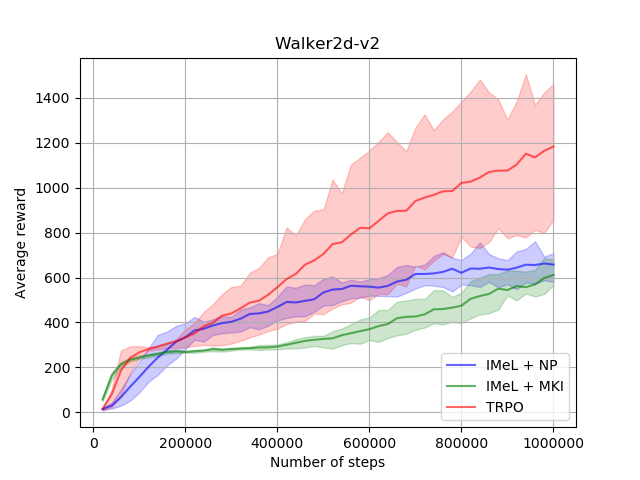}
\centering\subcaption{
$s \in \mathbb{R}^{17}$,
$a \in \mathbb{R}^{6}$
}
\end{minipage}%
\caption{
Comparison of IMeL with NPs, IMeL with MKI and TRPO in two benchmark problems.
IMeL shows superior performance in the low dimensional problem, while being competitive in the high dimensional one.
}
\label{fig:res_all}
\end{figure}

\section{Discussion}
\label{sec:discussion}

The proposed IMeL is a baseline RL algorithm based on interpolation of first order improvements of experiences in a RM.
IMeL is designed with the objective of delimiting the role of NN in the RL process. This objective is motivated by recent results which suggest that, as stated in \cite{ilyas2018deep}, ``\textit{the theoretical framework for deep RL algorithms is often unpredictive of phenomena arising in practice.}''

IMeL uses NNs as interpolators between procedurally updated policies in a reservoir, in a manner that could be extended to modeling environment dynamics and value learning. Therefore, we train a NN to measure the similarity between states in order to compute long-term forecasts by averaging experiences, rather than by encoding the problem structure in the NN parameters. We argue that this approach is more transparent to inspect whether the theoretical justifications of RL algorithms are met in practice. Moreover, RL tasks may call for specific inductive biases (e.g. partial-observability and/or states-from-pixels), in such cases the ability of NN to meet their requirements could be assessed separately from their performance in terms of returns.

In promising preliminary results, obtained for benchmark environments with a low-dimensional continuous-action space, we show that IMeL has comparable performance as TRPO (Fig.~\ref{fig:res_all}), both when employing NPs and MKI as interpolation method.
A properly tuned version of IMeL could be employed to evaluate the performance gains of more complex RL methods, such as the benefits of algorithmic advances and inductive biases.
These results are supported by additional tests reported in the Supplementary Material.

\bibliographystyle{abbrv}
\bibliography{main}

\appendix
\section{Additional results}
\begin{figure}[h!]
\begin{minipage}{0.5\linewidth}
\includegraphics[width=\linewidth]{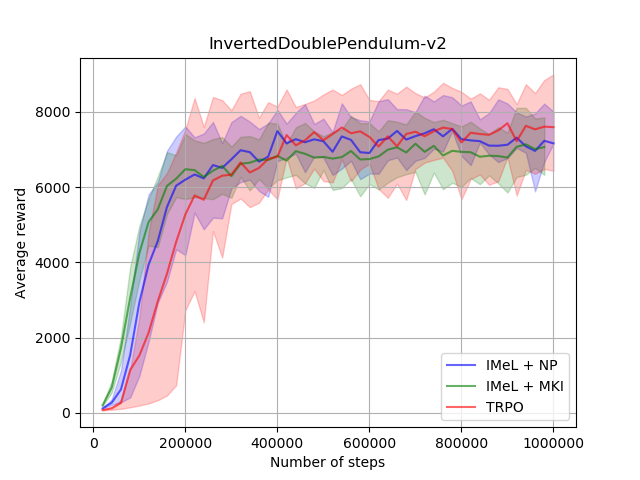}
\subcaption{$d_s = \mathbb{R}^{11}, \ d_a = \mathbb{R}$}\label{sub:1}
\end{minipage}%
\hfill
\begin{minipage}{0.5\linewidth}
\includegraphics[width=\linewidth]{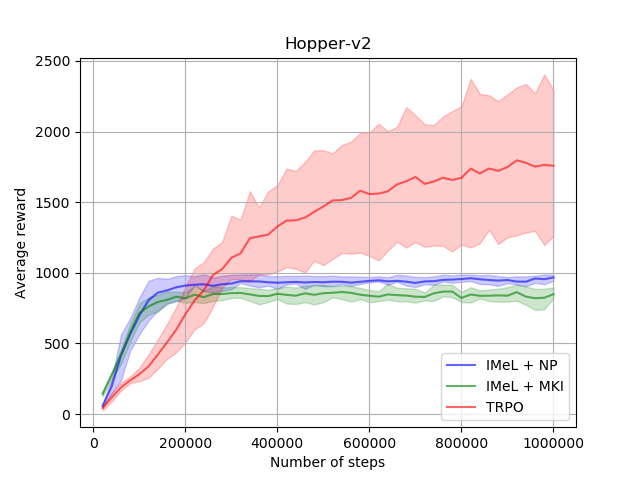}
\subcaption{$d_s = \mathbb{R}^{11}, \ d_a = \mathbb{R}^3$}\label{sub:2}
\end{minipage}%
\\
\begin{minipage}{0.5\linewidth}
\includegraphics[width=\linewidth]{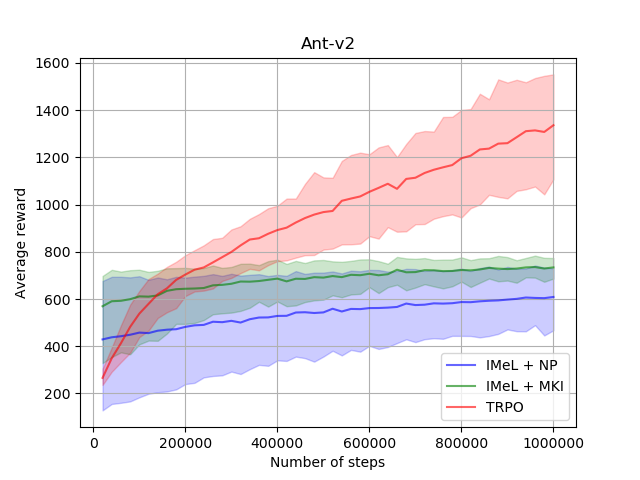}
\subcaption{$d_s = \mathbb{R}^{27}, \ d_a = \mathbb{R}^8$}\label{sub:3}
\end{minipage}%
\hfill
\begin{minipage}{0.5\linewidth}
\includegraphics[width=\linewidth]{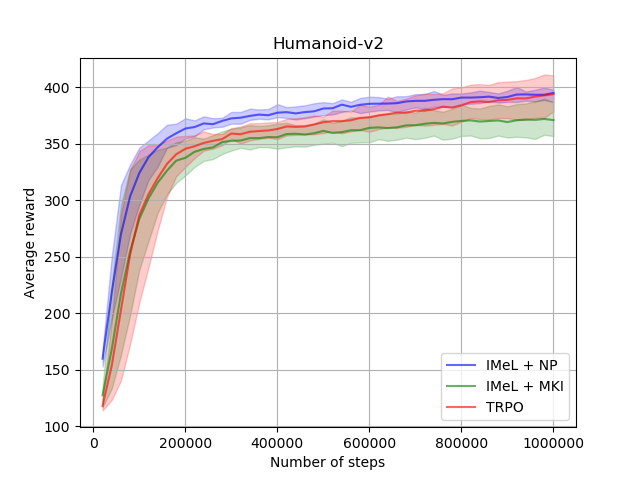}
\subcaption{$d_s = \mathbb{R}^{257}, \ d_a = \mathbb{R}^{17}$}\label{sub:4}
\end{minipage}%
\caption{
The average reward of IMeL equipped with Neural Process (NP), IMeL with Mean Kernel Interpolation (MKI), and Trust Region Policy Optimization algorithm (TRPO) is plotted with respect to the number of learning steps.
The shaded area around the mean reports the $20^{th}$ and $80^{th}$ percentiles of the reward.
We observe that IMeL is slightly superior compared to TRPO in both the Humanoid and the Inverted Double Pendulum environments.
In environments that are characterized by local optima, IMeL fails to recover from them and shows inferior performance compared to TRPO, demonstrating that the increased complexity in TRPO is necessary. 
In other environments, IMeL can be a faster, less sophisticated, computationally cheaper alternative.
}
\label{fig:res_add}
\end{figure}

IMeL was extensively evaluated in multiple environments from the Mujoco simulation engine.
In this section we provide the results on four environments, reported in Figure~\ref{fig:res_add}.

\section{Pseudocode for IMeL algorithm}
Here, we provide the pseudocode for the IMeL algorithm. 
The pseudocode of the IMeL algorithm equipped with NPs and MKI as supervised learning (SL) method is given respectively in Algorithm \ref{alg:imel_np} and Algorithm \ref{alg:imel_mki}.
To reduce the variance of the gradient estimate, IMeL computes the advantage function, as reported in Algorithm \ref{alg:adv}.
Eventually, the training procedure used to learn the improved policies distribution with MKI and NPs, is presented in Algorithm \ref{alg:train}.

\begin{algorithm}[H]
\caption{IMeL with Neural Processes}\label{alg:imel_np}

\begin{algorithmic}[1]
\STATE  
\textbf{Initial context} $\mathcal{C}^0 = \{ (s^0,a^0)_i \}_{i=1}^c$,  \ \ where: \ $s_{1:c}^0 \sim \mathcal{U}(S), \ a_{1:c}^0 \sim \mathcal{N}(\mu^0, \sigma^0)$

\textbf{SL model} \ $\model: \  (s_t, \mathcal{C}) \ \mapsto \ \pi(a_t|s_t)  \sim \mathcal{N}(\mu_t, \sigma_t)$

\FOR{$k=1$ {\bfseries to} $K$}
    \STATE $\operatorname{train} \big(\model, \mathcal{C}^{k-1} \big)$
    \FOR{$t=0$ {\bfseries to} $T$}
        \STATE $\mu_t^k, \sigma_t^k = \model(s_t,\ \mathcal{C}^{k-1})$  \hspace{.2cm} 
        \STATE Sample action $a_t$ from $\pi(a_t|s_t) \sim \mathcal{N}(\mu_t^k, \sigma_t^k )$ 
        \STATE Observe tuple $\mathcal{E}_{t}^k = \big(s_t, a_t, s_{t+1}, r_{t+1}, \mu^{k}_t, \sigma^{k}_t \big)$ 
    \ENDFOR
    \STATE $ A_{0:T} = \operatorname{estimateAdvantage} \big(s_{0:T}, r_{0:T}, \gamma \big)$
    \STATE $ \eta^k = \sqrt{\frac{2 \ T \ \epsilon}{ \sum_{t=0}^{T}{\frac{{A_t}^2(a_t - \mu_t^k)^2}{{\sigma^k}^6}}}}$\\
    \FOR{$t=1$ {\bfseries to} $T$}
        \STATE Update $\mu^*_t = \mu^{k}_t + \eta^k A_t \frac{a_t - \mu^{k}_t}{{\sigma^k}^2}$ \\
    \ENDFOR
    \STATE $\mathcal{C}^k = \mathcal{C}^{k-1} \ \ \cup \ \{(s, \mu^*)_t\}_{t=0}^T$\\
\ENDFOR
\end{algorithmic}
\end{algorithm}

\begin{algorithm}[H]
\caption{IMeL with Mean Kernel Interpolation}\label{alg:imel_mki}

\begin{algorithmic}[1]
\STATE  
\textbf{Initial context} $\mathcal{C}^0 = \{ (s^0,a^0)_i \}_{i=1}^c$,  \ \ where: \ $s_{1:c}^0 \sim \mathcal{U}(S), \ a_{1:c}^0 \sim \mathcal{N}(\mu^0, \sigma^0)$

\textbf{SL model} \ $\mathsf{MKI}^{\mathbf{w}}: \  (s_t, \mathcal{C}) \ \mapsto \ \mu_t$

\FOR{$k=1$ {\bfseries to} $K$}
    \STATE $\operatorname{train} \big( \mathsf{MKI}^{\mathbf{w}}$ , $\mathcal{C}^{k-1} \big)$ 
    \FOR{$t=0$ {\bfseries to} $T$}
        \STATE $\mu_t = \mathsf{MKI}^{\mathbf{w}}(s_t,\ \mathcal{C}^{k-1})$
        \STATE Sample action $a_t$ from $\pi(a_t|s_t) \sim \mathcal{N}(\mu_t^k, \sigma^k )$ 
        \STATE Observe tuple $\mathcal{E}_{t}^k = \big(s_t, a_t, s_{t+1}, r_{t+1}, \mu^{k}_t, \sigma^{k}_t \big)$ 
    \ENDFOR
    \STATE $ A_{0:T} = \operatorname{estimateAdvantage} \big(s_{0:T}, r_{0:T}, \gamma \big)$
    \STATE $ \eta^k = \sqrt{\frac{2 \ T \ \epsilon}{ \sum_{t=0}^{T}{\frac{{A_t}^2(a_t - \mu_t^k)^2}{{\sigma^k}^6}}}}$\\
    \FOR{$t=0$ {\bfseries to} $T$}
        \STATE Update $\mu^*_t = \mu^{k}_t + \eta^k A_t \frac{a_t - \mu^{k}_t}{{\sigma^k}^2}$ \\
        \STATE Compute $\sigma_t^* = \sigma_t^k + \eta^k \ \frac{(a_t - \mu^k_t)^2 - {\sigma^k_t}^2}{{\sigma^k_t}^3}$ \\
        \STATE Update $\sigma^{k+1} = \frac{1}{T} \sum_{t=0}^{T} \sigma_t^*$
    \ENDFOR
    \STATE $\mathcal{C}^k = \mathcal{C}^{k-1} \ \ \cup \ \{(s, \mu^*)_t\}_{t=0}^T$\\
\ENDFOR
\end{algorithmic}
\end{algorithm}

\begin{algorithm}
\caption{Advantage estimate}\label{alg:adv}
\begin{algorithmic}
    \STATE
    $\operatorname{estimateAdvantage}\big(s_{0:T}, r_{0:T}, \gamma \big)$:
    \FOR{$t=0$ {\bfseries to} $T$}
    \STATE $q_t=\sum_{l=1}^{T'} \gamma^{l}r_{t+l}$
    \STATE Estimate $v^{NN}_t = V^{\mathbf{w}}(s_t)$    \IF{gae} \STATE $A_t = GAE^{\lambda, \gamma}(v^{NN}_t, q_t)$
    \ELSE \STATE $A_t = q_t - v^{NN}_t$
    \ENDIF
    \ENDFOR
    \STATE \textbf{return} $A_{0:T}$
\end{algorithmic}
\end{algorithm}

\begin{algorithm}[H]
\begin{algorithmic}
\label{SLtraining}

\caption{Leave-one-out training}\label{alg:train}
\STATE {\bfseries Input:} Replay Memory (RM) containing $M$ episodes: $ \mathcal{C} = \{\{(s, \mu^*)^m_t\}_{t=0}^{T_m}\}_{m=0}^M$
\STATE 
$\operatorname{train} \big(\model, \ \mathcal{C} \big)$:
\FOR {$e=0$ {\bfseries to} $epochs$}
  \FOR {$m=0$ {\bfseries to}  $M$}
    \STATE Remove episode $m$ from RM: \  $\mathcal{C}^{-m} = \ $ $\mathcal{C}$ $\setminus \ \{(s, \mu^*)^m_t\}_{t=0}^{T_m}$
    \STATE Target set: \  $\mathcal{T}^m = \ \{(s, \mu^*)^m_t\}_{t=0}^{T_m}$ 
    \STATE Predict: \  $\pi_{1:T_m} = \model(s_{1:T_m}, \mathcal{C}^{-m})$ 
    \STATE Compute loss: \  $\mathcal{L}(\pi_{1:T_m}, \mu^*_{1:T_m})$
    \STATE Optimize model: \  $\mathbf{w}'$ $ = $ $\mathbf{w}$ $ - \alpha \frac{\partial \mathcal{L}}{\partial \mathbf{w}}$
  \ENDFOR
\ENDFOR
\end{algorithmic}
\end{algorithm}

\end{document}